\documentclass[twoside,11pt]{article}

\usepackage{jmlr2e}
\usepackage{pgfplots}
\pgfplotsset{compat=1.16}
\usepgfplotslibrary{groupplots}
\usepackage{amsmath}
\usepackage{bm}
\usepackage{tikzscale}
\usepackage{caption}
\usepackage{subcaption}

\newcommand{\minmax}{Min Max~}

\ShortHeadings{ICML 2020 Workshop on Real World Experiment Design and Active Learning}{Workshop on Real World Experiment Design and Active Learning}
\firstpageno{1}

\begin{document}

\title{Bayesian Optimization for Min Max Optimization}

\author{\name Dorina Weichert \email dorina.weichert@iais.fraunhofer.de \\
       \addr Fraunhofer Center for Machine Learning \\
       Fraunhofer Institute for Intelligent Analysis and Information Systems IAIS \\
       Schloss Birlinghoven,
       53757 Sankt Augustin, Germany \\ 
       \AND
       \name Alexander Kister \email alexander.kister@iais.fraunhofer.de \\
       \addr Fraunhofer Institute for Intelligent Analysis and Information Systems IAIS\\
       Schloss Birlinghoven,
       53757 Sankt Augustin, Germany}

\editor{}

\maketitle

\begin{abstract}
A solution that is only reliable under favourable conditions is hardly a safe solution. 
Min Max Optimization is an approach that returns optima that are robust against worst case conditions.
We propose algorithms that perform Min Max Optimization in a setting where the function that should be optimized is not known a priori and hence has to be learned by experiments. Therefore we extend the Bayesian Optimization setting, which is tailored to maximization problems, to Min Max Optimization problems. While related work extends the two acquisition functions Expected Improvement and Gaussian Process Upper Confidence Bound; we extend the two acquisition functions Entropy Search and Knowledge Gradient. 
These acquisition functions are able to gain knowledge about the optimum instead of just looking for points that are supposed to be optimal. In our evaluation we show that these acquisition functions allow for better solutions - converging faster to the optimum than the benchmark settings.
\end{abstract}


\section{Introduction}%
Only in the lab, during development, engineers have control over the environment in which their artifact has to work. 
When the artifact is used,
these environmental conditions are out of control. So to guarantee a certain level of performance, engineers have to find a solution that is robust against changes in the environment.
For this optimization we formally assume the artifacts' performance is given by a function $f:\Theta \times Z \to \mathbb{R}$, where $\Theta$ is the space of controllable parameters and $Z$ is the space of environment conditions, and use the \minmax Optimization given by:
\begin{equation}
    \min_{\theta \in \Theta} \max_{\zeta \in Z} f(\theta, \zeta).
\end{equation}
In this paper, we will study this problem with a continuous set of controllable parameters $\Theta$ and a discrete set of uncontrollable parameters $Z$. 

In many practical situations the performance metric, $f(\theta,\zeta)$, is unknown and has to be treated as a black box. Bayesian Optimization is an established iterative framework for optimization of black box functions, that is - due to its sample efficiency - particularly useful when function evaluations are costly (\cite{Shahriari2016}). In each iteration, Bayesian Optimization suggests a new location $(\bm{\theta}_{n+1},\bm{\zeta}_{n+1})$. After evaluating the black box function at this location, yielding the observation $y$, the existing data set of observations $\mathcal{D}_n$ is extended with this observation $\mathcal{D}_{n+1}= \mathcal{D}_{n}\cup \{((\theta_{n+1},\zeta_{n+1}),y)\}$. Typically Bayesian Optimization uses a Gaussian Process (\cite{Rasmussen2006}) as a surrogate model and an acquisition function for suggesting the next experiment to be performed (or, in other words, the next parameter to be evaluated), (\cite{Srinivas2010, Hennig2012, Frazier2018, Shahriari2016}).

A distinguishing feature between acquisition functions are the properties a returned candidate is expected to have. The Gaussian Process Upper Confidence Bound (GP-UCB) (\cite{Srinivas2010}) and Expected Improvement (\cite{Jones1998}) return candidates that are potentially the optimum, while Entropy Search (\cite{Hennig2012}) and Knowledge Gradient (\cite{Frazier2018,DBLP:journals/informs/FrazierPD09}) return candidates that increase the information about the optimum. So, as the information can increase also by by the exclusion of candidates, these latter acquisition functions explicitly encourage to evaluate non optimal points.

\textbf{Related Work} The existing approaches for extending the Bayesian Optimization framework to the \minmax problem usually pick an acquisition function that was designed to work for maximization problems and alter it such that it also works for the \minmax problem. The literature is dominated by approaches that extend GP-UCB (e.g., \cite{Sessa2020, Bogunovic, Wabersich2015}) or the Expected Improvement (\cite{UrRehman2015},\cite{Marzat2016}). They use two acquisition functions: one to find the candidate for the controllable parameter and one to find the uncontrollable parameter that is a good candidate for the worst case match for the controllable parameter candidate. 
The GP-UCB approaches of \cite{Sessa2020} and \cite{Bogunovic} tackle problems that cover the \minmax problem as a special case.

Our approach deals with the adaption of two 
acquisition functions: Entropy Search and Knowledge Gradient. They were already adapted to the case when robustness is not required with respect to the worst case but with respect to the mean: \cite{Frohlich2020} adapted the Entropy Search while \cite{Toscano2018} adapted the Knowledge Gradient. While, for the Knowledge Gradient, the necessary adaptions are similar for both kinds of robustness, they differ strongly when it comes to Entropy Search. 

In contrast to the GP-UCB and Expected Improvement approaches, the focus of Entropy Search and Knowledge Gradient on information improvement allows us to use only one acquisition function. 
Our hypothesis is that the empirically observed good performance of Entropy Search and Knowledge Gradient for the maximization problem are inherited by our adaptions to the \minmax Optimization.

\textbf{Contributions} In short, our contributions are: (i) the adaption of the Entropy Search and Knowledge Gradient acquisition functions for the \minmax problem, (ii) the demonstration of their efficiency in comparison to Thompson Sampling (\cite{Thompson1933}) and the algorithm of \cite{Wabersich2015} (on synthetic problems), (iii) the discussion of the results, 
showing the advantageous behaviour of our adaptions 
and (iv) an outline of our future work.

\section{Bayesian Optimization for Min Max Optimization}
\label{sec:adaptions}
We tailor the two acquisition functions, Entropy Search (\cite{Hennig2012}) and Knowledge Gradient (\cite{Frazier2018,DBLP:journals/informs/FrazierPD09}), so that they are applicable for searching the \minmax point.

 \textbf{Entropy Search}
In Entropy Search, we seek candidates that improve the knowledge about the location of the optimum. Given the set of observations $\mathcal{D}_n$, the Entropy Search acquisition function is defined by:
$
    \alpha_{\text{ES}}((\bm{\theta},\bm{\zeta}); \mathcal{D}_n) := \mathbb{E}\left[\eta^\star_n - \eta^\star_{n+1} \right | (\bm{\theta}_{n+1},\bm{\zeta}_{n+1}) = (\bm{\theta},\bm{\zeta})] ,
$
where $\eta^\star_n$ is the entropy of a distribution $p_{\text{opt}}$ derived from the current Gaussian Process surrogate model that represents our knowledge about the location of the optimum and $\eta^\star_{n+1}=\eta^\star_{n+1}(y)$ is the same quantity derived from the Gaussian Process surrogate model that was updated with the fictive observation $y$ at location $(\bm{\theta},\bm{\zeta})$. The expectation is taken with respect to the measure for the fictive observation $y$ induced by the (not updated) current surrogate model.
The distribution $p_{\text{opt}}$ is given by
$
p_{\text{opt}}((\bm{\theta},\bm{\zeta}))=p((\bm{\theta},\bm{\zeta})=(\bm{\theta}^\star,\bm{\zeta}^\star)),  
$
where $p$ is the measure of the Gaussian Process surrogate model and $(\bm{\theta}^\star,\bm{\zeta}^\star)$ is the searched optimum. In our context, this is the \minmax point and in the original Entropy Search setting it is the global maximum.

In the original maximization setting there is no closed form expression for the distribution and instead two approximations are used: first, the uncountable search space is replaced by a finite set of representative points. Second, Expectation Propagation (EP) (\cite{Minka2013}) is used for estimating the distribution over the optimum within these representative points. We follow the first approximation and introduce the representative controllable parameters $\bm{\theta}_{1},\bm{\theta}_{2},....,\bm{\theta}_{N}$. While following the second approximation we encountered the obstacle that in the \minmax setting the distribution does not have a nice multiplicative decomposition into terms that depend on maximally two locations of the Gaussian Process (as it is needed for EP). Instead, we express: $p_{\text{opt}}((\bm{\theta},\bm{\zeta}))=p_{\text{Min Max}}\left(\left(\bm{\theta}_{i^\star},\bm{\zeta}^\star\right)\right)$ as
\begin{equation}\label{eq:p_min_max_direct}
\int 
\prod_{\substack{\bm{\zeta} \in Z \\ \bm{\zeta}^\star \neq \bm{\zeta}}} H\left[f\left(\bm{\theta}_{i^\star},\bm{\zeta}^\star\right) - f\left(\bm{\theta}_{i^\star},\bm{\zeta}\right)\right] 
\prod_{i\in \left\{1,...,N\right\} \setminus  i^\star} \left(1-\prod_{\substack{\bm{\zeta} \in Z \\ \bm{\zeta}^\star \neq \bm{\zeta}}} H \left[ f \left( \bm{\theta}_{i^\star}, \bm{\zeta}^ \star \right) - f \left( \bm{\theta}_{i}, \bm{\zeta} \right) \right] \right) p\left( f \right) d f, 
\end{equation}
where $H$ is the heavy side function and $p$ is the measure of the Gaussian Process. 
To derive this expression we used that stating the point $(\bm{\theta}^\star,\bm{\zeta}^\star)$ is the \minmax point of function $f$ is equivalent to the following two statements: 
\begin{itemize}
    \item For the worst case optimal controllable parameter $\bm{\theta}_{i^\star}$, the function value $f(\bm{\theta}_{i^\star},\bm{\zeta}^\star)$ is the worst case among all uncontrollable parameter; in short: $f(\bm{\theta}_{i^\star},\bm{\zeta}^\star)\geq f(\bm{\theta}_{i^\star},\bm{\zeta})$ for all $\bm{\zeta} \in Z$, 
    \item For all other controllable parameter settings $\bm{\theta}_{i}$ for $i\in \left\{1,...,N\right\} \setminus  i^\star$, the function value $f(\bm{\theta}_{i^\star},\bm{\zeta}^\star)$ is exceeded for at least one uncontrollable parameter setting; in short $f(\bm{\theta}_{i^\star},\bm{\zeta}^\star)\leq f(\bm{\theta}_{i},\bm{\zeta})$ for at least one $\bm{\zeta} \in Z$.
\end{itemize}
Our workaround for the sum that emerges after multiplying out the second product in formula~\ref{eq:p_min_max_direct} is to condition on the argmax function $g$, this is the function that maps each controllable parameter setting $\bm{\theta}$ to the worst case uncontrollable parameter setting. We consider the conditional probability of $\bm{\theta}_{i^\star}$ being the minimizer of the worst case function:
$
    P(\bm{\theta}_{i^\star} \text{ is optimal}~|~g~ \text{is the argmax function}).~
$
Up to a normalization constant this conditional probability is given by
\begin{equation}\label{eq:p_min_max_given_arg_max}
\int \prod_{i =1}^{N}\prod_{\substack{\bm{\zeta} \in Z \\ \bm{\zeta}^\star \neq \bm{\zeta}}}H\left[f\left(\bm{\theta}_{i},g\left(\bm{\theta}_{i}\right)\right)-f\left(\bm{\theta}_{i},\bm{\zeta}\right)\right]
\prod_{i\in \{1,...,N\} \setminus  i^\star} H\left[f\left(\bm{\theta}_{i},g\left(\bm{\theta}_{i}\right)\right)-f\left(\bm{\theta}_{i^{*}},g\left(\bm{\theta}_{i^{*}}\right)\right)\right]
p\left(f\right) df,
\end{equation}
where the first product (over the index $i$) under the integral is the indicator for the function $g$ to actually be the argmax function and the second product is the indicator for $\bm{\theta}_{i^\star}$ to be the minimizer of the worst case function $f(\bm{\theta}_{i},g(\bm{\theta}_{i}))$. To estimate (the unconditional) probability of $\bm{\theta}_{i^\star}$ being the minimizer of the worst case function, 
we sample argmax functions $g_{1},g_{2},...,g_{M}$ from the Gaussian Process and take the mean of the resulting conditional probabilities.

As the integrand in formula~\ref{eq:p_min_max_given_arg_max} is a product of terms that depend on maximally two locations of a Gaussian Process, we can reuse large parts of existing Entropy Search implementations: our implementation is based on the GPyOpt package (\cite{gpyopt2016}).

\textbf{Knowledge Gradient}
The Knowledge Gradient acquisition function $\alpha_{\text{KG}}$ (\cite{DBLP:journals/informs/FrazierPD09,Frazier2018}) reflects how strongly the optimum of the mean of the surrogate model is influenced by a function evaluation at a given location. 

The acquisition function is defined analogously to the one for Entropy Search. The entropies, $\eta^\star_n$ and $\eta^\star_{n+1}(y)$, are replaced by the means, $\mu^\star_n$ and $\mu^\star_{n+1}(y)$, of the corresponding Gaussian Processes:
$
    \alpha_{\text{KG}}((\bm{\theta},\bm{\zeta}); \mathcal{D}_n) := \mathbb{E}\left[\mu^\star_n - \mu^\star_{n+1} \right | (\bm{\theta}_{n+1},\bm{\zeta}_{n+1}) = (\bm{\theta},\bm{\zeta})] ,
$

For maximization, an implementation is described in~\cite{Frazier2018} Algorithm 2. Adopting this approach, that uses Monte Carlo estimates for the expectation and grid search for maximising the acquisition function, is straight forward. 
The more scalable approach, that uses stochastic gradient descent for finding the maximum of the acquisition function, as implemented in Algorithm 3 and 4 in \cite{Frazier2018}, is left for future work.

\section{Experiments}
\label{sec:experiments}
\begin{figure}
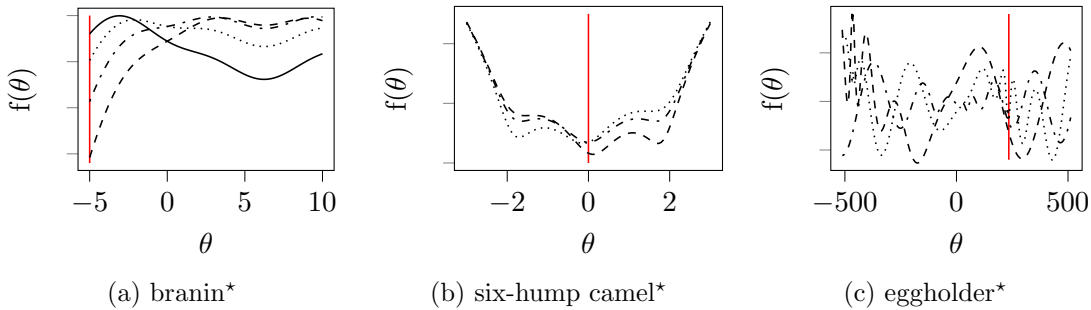

    \begin{subfigure}[t]{0.32\textwidth}
        \centering
        \includegraphics[width=0.95\textwidth, height=0.7\textwidth]{testproblem1.tikz}
        \subcaption{branin$^\star$}
    \end{subfigure}
    \begin{subfigure}[t]{0.32\textwidth}
        \centering
        \includegraphics[width=0.95\textwidth, height=0.7\textwidth]{testproblem2.tikz}
        \subcaption{six-hump camel$^\star$}
    \end{subfigure}
    \begin{subfigure}[t]{0.32\textwidth}
        \centering
        \includegraphics[width=0.95\textwidth, height=0.7\textwidth]{testproblem3.tikz}
        \subcaption{eggholder$^\star$}
    \end{subfigure}
\caption{The three different test problems are adaptions of the functions branin, six-hump camel and eggholder. The vertical line indicates the \minmax location. In branin$^\star$, the \minmax point is at the boundary and the argmax function is nearly constant. In six-hump camel$^\star$ the \minmax point is at a non differentiable point of the argmax functions. In eggholder$^\star$, the optimum lies in a local optimum of one of the functions and the argmax changes frequently. }
\label{fig:testproblems}
\end{figure}

To test our approaches, we adapt three synthetic two-dimensional problems, namely the branin, the six-hump camel and the eggholder function\footnote{\texttt{https://www.sfu.ca/$\sim$ssurjano/optimization.html}}.
Our test cases, visualized in Figure~\ref{fig:testproblems}, are representative for three problems: the robust optimum is at the boundary, at a not differentiable and at a differentiable location of the worst case function. 
We benchmark our methods\footnote{\texttt{https://github.com/fraunhofer-iais/MinMaxOpt}} against Thompson Sampling (\cite{Thompson1933}) and the GP-UCB approach of \cite{Wabersich2015}. 

Further details about our experimental setup and the benchmark implementations can be found in appendix~A and appendix~B.

\begin{figure}
    \centering
    \includegraphics{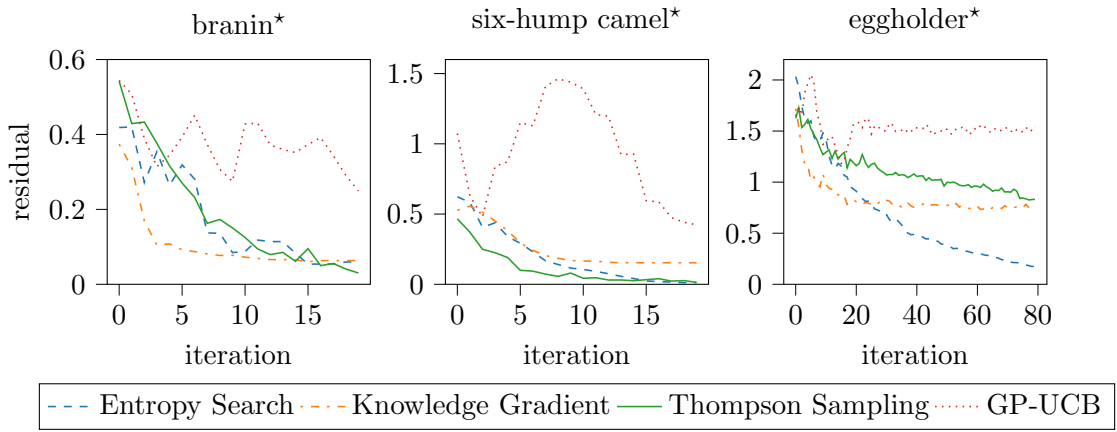}
    \caption{Mean residuals of the different algorithms on the three test problems.}\label{fig:residuals}
\end{figure}

We run 100 trials with randomized initializations of (in total) 5 points and measure the mean absolute residuals (the mean absolute difference of the current function value at the estimated \minmax location and the function value at the real \minmax location), see Figure~\ref{fig:residuals}. For all test cases, Entropy Search, Knowledge Gradient and Thompson Sampling show superior performance over the benchmark of \cite{Wabersich2015}. 
The benchmark's unfavourable behaviour is caused by the deteriorations due to maximization in the inner loop that could not be counterbalanced by the improvement due to minimization in the outer loop.
This is particularly noticeable for the branin$^\star$ and the camel$^\star$ problem, as the oscillation is especially large in the first iterations, when the minimizing loop lacks a sufficient amount of data. For eggholder$^\star$, the performance in the first iterations is comparable to the other acquisition functions, but the algorithm of \cite{Wabersich2015} tends to get stuck in a local minimum. 
Knowledge Gradient shows a fast convergence in the first iterations, but sticks to local optima as well, as can be seen for the eggholder$^\star$ case. This is due to a lack of exploration, as the Knowledge Gradient only concentrates on the difference of the mean function values.
On the contrary, Thompson Sampling, that performs well on branin$^\star$ and camel$^\star$, explores too heavily on the eggholder$^\star$ problem. This is due to the short (and fixed) lengthscales of the underlying Gaussian Process, as their is a high variance in the samples drawn from it, when the amount of training data is low. 
Entropy Search is comparable to Thompson Sampling in the smooth problems (branin$^\star$ and camel$^\star$). But it shows its superior performance on the eggholder$^\star$ problem, as it does not explore as aggressively as Thompson Sampling, but exploits the problem structure when estimating the probability of being the \minmax location.
Further analyses (e.g., standard deviations, influence of algorithm parameters in Entropy search) are provided in appendix~C.

\section{Conclusion and Future Work}
\label{sec:conclusion}
We extended two existing acquisition functions - Entropy Search and Knowledge Gradient - such that they are applicable for solving the \minmax problem. We compared and benchmarked the algorithms on three representative problems. We could show the comparable or advantageous performance of our approaches against two existing benchmarks, namely Thompson Sampling (\cite{Thompson1933}) and \cite{Wabersich2015}. 
The Knowledge Gradient acquisition function shows a very fast convergence during the first iterations but might get stuck in local minima. Entropy search converges slower, but it comes close to the optimum even for the eggholder problem in an acceptable number of iterations. Regarding our hypothesis in the beginning, we showed that the empirically shown good performance of the acquisition functions Knowledge Gradient and Entropy Search for maximization is transferable to \minmax Optimization.

\textbf{Future Work} 
There are two directions for future work: enhancements to the adaption of the Entropy Search acquisition function and to the experimental sections. For the adoption of the Entropy Search acquisition function, \cite{Gessner2020} enables us to directly approximate Equation~\ref{eq:p_min_max_direct} with the Expectation Propagation algorithm, instead of enforcing a product representation as we are doing currently by conditioning on the argmax function.

Furthermore, Max-value Entropy Search (\cite{DBLP:conf/icml/WangJ17}) allows to apply that the \minmax value is the maximal value that is, for each controllable parameter setting, exceeded (including matched) for at least one setting of the uncontrollable parameter. 
This alternative approach has the potential to be more accurate and easier to compute as it was seen where the goal was maximization (\cite{DBLP:conf/icml/WangJ17}), where it was robustness with respect to the mean (\cite{Frohlich2020}), and where it was Pareto optimality (\cite{DBLP:conf/nips/BelakariaDD19}).

We plan to perform tests on higher dimensional test settings and to benchmark against other state-of-the-art approaches such as (\cite{Sessa2020}).
Additionally, as it is easy to extend our approaches to the case where the space of uncontrollable parameters $Z$ is not finite, we will perform experiments to test its performance for this setting.
\acks{This work was jointly developed by the Fraunhofer Center for Machine Learning within the Fraunhofer Cluster for Cognitive Internet Technologies and the Fraunhofer Lighthouse projects ML4P and SWAP.}
\newpage

\appendix
\section*{Appendix A. Test Setup}
\label{app:TestSetup}
During the optimization, we use a Gaussian Process with an automatic relevance determination squared-exponential covariance function and zero mean function with fixed hyperparameters (signal variance $\sigma_v^2$, lengthscales $l$ and noise variance $\sigma_n^2$), to avoid disturbances of the analysis due to wrongly estimated hyperparameters.

Furthermore, we applied the Gaussian Process model to scaled versions of the functions: the input locations of the functions are fit to the bounding box $\left[0, 1\right]^2$ and the output values of the functions are normalized to zero mean and variance 1. A summary of the test setups is provided in Table~\ref{tab:testproblems}. 

We developed our code with the use of the python packages GPyOpt (\cite{gpyopt2016}) and BoTorch (\cite{balandat2019botorch}).
\begin{table}[t]
    \centering
    \begin{tabular}{c|c|c|c}
         & branin$^\star$ & six-hump camel$^\star$ & eggholder$^\star$  \\ \hline
         description & negated branin & six-hump camel$^\dagger$ & eggholder \\
         fixed slices & \{0, 4, 8, 12\} & \{-0.9, 0, 1\} & \{-512, 0, 185\}\\
         min-max location & (-5, 12) & (0, 0), (0, 1) & (234.647671, 185) \\
         $\left(\sigma_n, \sigma_v, l\right)$ & (0.001, 1, (0.2, 0.4)) & (0.001, 0.5, (0.2, 0.2)) & (0.001, 1, (0.09, 0.09)) 
         \end{tabular}
    \caption{The test problems are adaptions of the traditional branin, six-hump camel and eggholder function. To match our \minmax problem, we treat the first dimension as controllable parameters $\bm{\theta}$ and the second as uncontrollable parameters $\bm{\zeta}$. We fix a set of parameters in the second dimension, producing multiple slices. Additionally, we make small adaptions to construct interesting problems for \minmax Optimization. The parameters $\left(\sigma_n, \sigma_v, l\right)$ are the hyperparameters we used for the Gaussian Process. $^\dagger$:  restricted to $[-3, 3] \times [-2, 2]$ and transformed by $\log(f(\bm{\theta}, \bm{\zeta}) + 2)$.}
    \label{tab:testproblems}
\end{table}

\section*{Appendix B. Benchmarks}
\label{app:BenchmarkSetup}
For Thompson Sampling, a sample of a Gaussian Process is drawn in every iteration and its optimum location (here: the Min Max) used for the next evaluation. As we use a discretization to find the \minmax location of the sample (due to discontinuous partial first derivatives of the worst case function we cannot use traditional gradient based optimizers), 
we also use this discretization for sampling from the Gaussian Process, avoiding expensive operations like spectral sampling (\cite{lazaro-gredilla10a}).

The algorithm of \cite{Wabersich2015} stays in the nested setting of the \minmax problem, resulting in an outer optimization loop for minimizing and an inner for maximimizing given the current candidate for the minimum. Here, GP-UCB with a tailored exploration-exploitation-tradeoff parameter $\beta$ for the current optimization state, favouring exploration at the beginning of the optimization and exploitation at its end, is used.

\section*{Appendix C. Further results}
\label{app:further results}

\begin{figure}
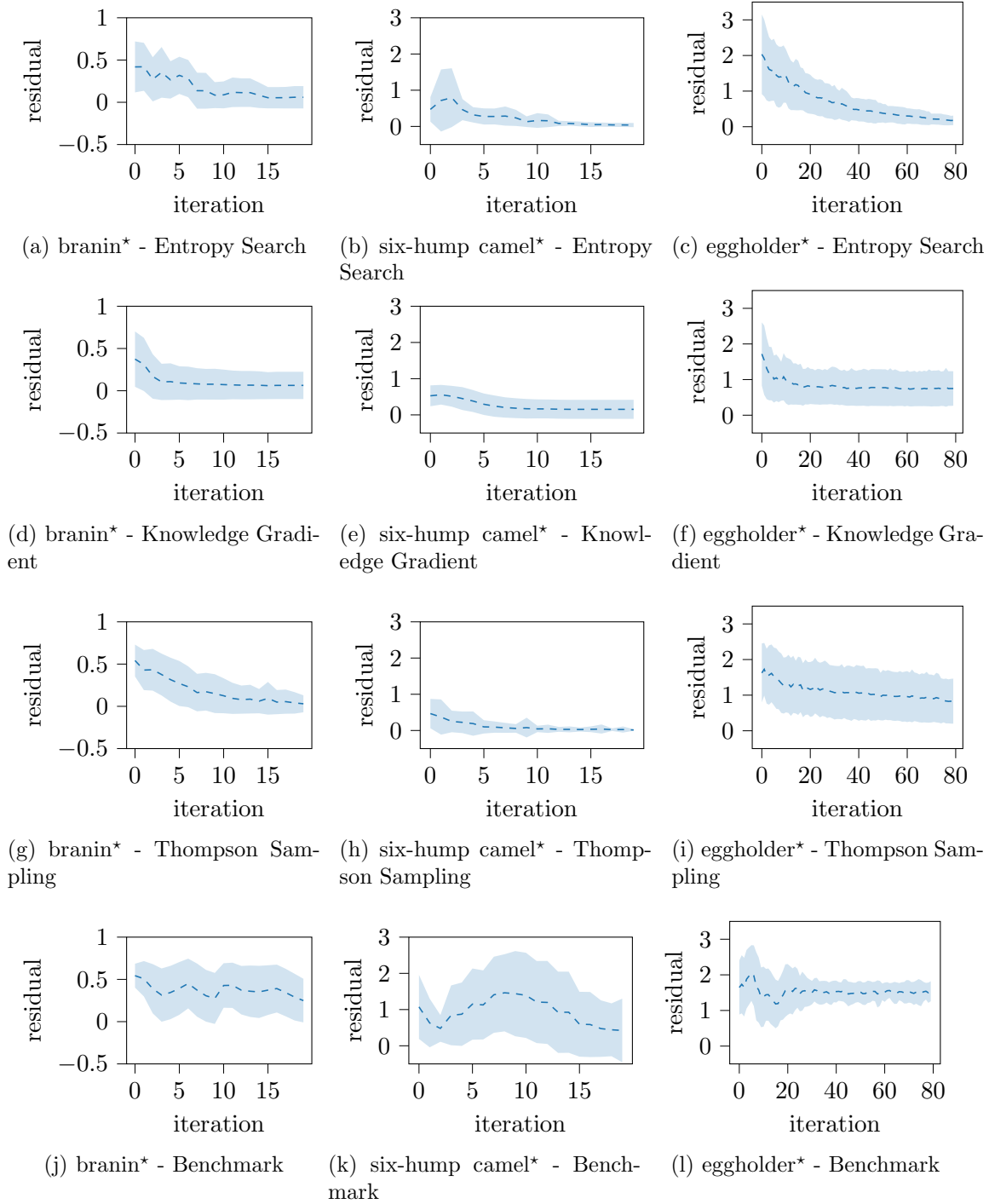

    \begin{subfigure}[t]{0.32\textwidth}
        \centering
        \includegraphics[width=0.95\textwidth, height=0.7\textwidth]{branin_entropy.tikz}
        \subcaption{branin$^\star$ - Entropy Search}
    \end{subfigure}
    \begin{subfigure}[t]{0.32\textwidth}
        \centering
        \includegraphics[width=0.95\textwidth, height=0.7\textwidth]{camel_entropy.tikz}
        \subcaption{six-hump camel$^\star$ - Entropy Search}
    \end{subfigure}
    \begin{subfigure}[t]{0.32\textwidth}
        \centering
        \includegraphics[width=0.95\textwidth, height=0.7\textwidth]{eggholder_entropy.tikz}
        \subcaption{eggholder$^\star$ - Entropy Search}
    \end{subfigure}
    \bigskip
        \begin{subfigure}[t]{0.32\textwidth}
        \centering
        \includegraphics[width=0.95\textwidth, height=0.7\textwidth]{branin_KGgrid.tikz}
        \subcaption{branin$^\star$ - Knowledge Gradient}
    \end{subfigure}
    \begin{subfigure}[t]{0.32\textwidth}
        \centering
        \includegraphics[width=0.95\textwidth, height=0.7\textwidth]{camel_KGgrid.tikz}
        \subcaption{six-hump camel$^\star$ - Knowledge Gradient}
    \end{subfigure}
    \begin{subfigure}[t]{0.32\textwidth}
        \centering
        \includegraphics[width=0.95\textwidth, height=0.7\textwidth]{eggholder_KGgrid.tikz}
        \subcaption{eggholder$^\star$ - Knowledge Gradient}
    \end{subfigure}
    \bigskip
    \begin{subfigure}[t]{0.32\textwidth}
        \centering
        \includegraphics[width=0.95\textwidth, height=0.7\textwidth]{branin_Thompson.tikz}
        \subcaption{branin$^\star$ - Thompson Sampling}
    \end{subfigure}
    \begin{subfigure}[t]{0.32\textwidth}
        \centering
        \includegraphics[width=0.95\textwidth, height=0.7\textwidth]{camel_Thompson.tikz}
        \subcaption{six-hump camel$^\star$ - Thompson Sampling}
    \end{subfigure}
    \begin{subfigure}[t]{0.32\textwidth}
        \centering
        \includegraphics[width=0.95\textwidth, height=0.7\textwidth]{eggholder_Thompson.tikz}
        \subcaption{eggholder$^\star$ - Thompson Sampling}
    \end{subfigure}
    \bigskip
    \begin{subfigure}[t]{0.32\textwidth}
        \centering
        \includegraphics[width=0.95\textwidth, height=0.7\textwidth]{branin_Benchmark.tikz}
        \subcaption{branin$^\star$ - Benchmark}
    \end{subfigure}
    \begin{subfigure}[t]{0.32\textwidth}
        \centering
        \includegraphics[width=0.95\textwidth, height=0.7\textwidth]{camel_Benchmark.tikz}
        \subcaption{six-hump camel$^\star$ - Benchmark}
    \end{subfigure}
    \begin{subfigure}[t]{0.32\textwidth}
        \centering
        \includegraphics[width=0.95\textwidth, height=0.7\textwidth]{eggholder_Benchmark.tikz}
        \subcaption{eggholder$^\star$ - Benchmark}
    \end{subfigure}
\caption{Standard deviations (one standard deviation interval is coloured) and means of residuals of the acquisition functions on the test problems. The benchmark of \cite{Wabersich2015} is simply called `"Benchmark"' here.}
\label{fig:stds}
\end{figure}
For the standard deviations of the acquisition functions on the test cases, see Figure~\ref{fig:stds}. The overall large size of standard deviations is due to the low number of samples for the initialization. In the worst case, all 5 samples have the same coordinate for the second axis, resulting in a long exploration phase. The standard deviations of the information-based acquisition functions Entropy Search and Thompson sampling shrink with a higher number of iterations, as the algorithm gains further knowledge about the location of the optimum. 

\begin{figure}
    \centering
    \includegraphics[width=0.6\textwidth, height=0.4\textwidth]{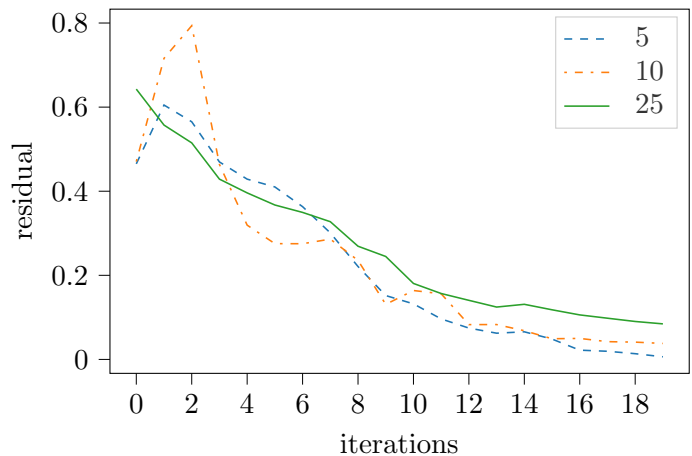}
    \caption{Residuals on the camel problem for different numbers of representative points and a fixed number of 10 argmax samples.}
    \label{fig:my_label}
\end{figure}
For the Entropy Search algorithm the number of representative points has a smoothing effect on the performance, see Figure~\ref{fig:my_label}. The higher the number of samples, the smoother the convergence plot; the speed of convergence is not effected (small differences are due to the high standard deviations of our experiments). 

\vskip 0.2in
\newpage
\bibliography{main_paper}

\end{document}